\documentclass[nohyperref]{article}

\usepackage{microtype}
\usepackage{graphicx}
\usepackage{xspace}
\usepackage{wrapfig}
\usepackage{booktabs} % for professional tables
\usepackage[position=top]{subfig}
\usepackage[pagebackref=true, colorlinks=true,urlcolor=blue,citecolor=black,linkcolor=blue,anchorcolor=black]{hyperref}

\usepackage[accepted]{icml2023}

\usepackage{amsmath}
\usepackage{amssymb}
\usepackage{mathtools}
\usepackage{amsthm}

\usepackage[capitalize,noabbrev]{cleveref}
\usepackage[shortlabels]{enumitem}
\setlist[enumerate, 1]{1\textsuperscript{o}}
\theoremstyle{plain}

\theoremstyle{definition}

\theoremstyle{remark}

\usepackage[textsize=tiny]{todonotes}

\renewcommand*{\backref}[1]{}
\renewcommand*{\backrefalt}[4]{%
    \ifcase #1%
          \or Cited on page~#2.%
    \else Cited on pages~#2.%
\fi%
}

\icmltitlerunning{Investigating the role of model-based learning in exploration and transfer}

\begin{document}

\twocolumn[
\icmltitle{Investigating the role of model-based learning in exploration and transfer}



\icmlsetsymbol{equal}{*}
\icmlsetsymbol{equal2}{†}

\begin{icmlauthorlist}
\icmlauthor{Jacob Walker}{equal,dm}
\icmlauthor{Eszter Vértes}{equal,dm}
\icmlauthor{Yazhe Li}{equal,dm}
\icmlauthor{Gabriel Dulac-Arnold}{gb}
\icmlauthor{Ankesh Anand}{dm}
\icmlauthor{Théophane Weber}{equal2,dm}
\icmlauthor{Jessica B. Hamrick}{equal2,dm}
\end{icmlauthorlist}

\icmlaffiliation{dm}{DeepMind, London, UK}
\icmlaffiliation{gb}{Google Research}

\icmlcorrespondingauthor{Jacob Walker}{jcwalker@deepmind.com}
\icmlcorrespondingauthor{Eszter Vértes}{evertes@deepmind.com}
\icmlcorrespondingauthor{Yazhe Li}{yazhe@deepmind.com}

\icmlkeywords{Machine Learning, ICML}

\vskip 0.3in
]



\printAffiliationsAndNotice{\icmlEqualContribution} % otherwise use the standard text.

\begin{abstract}
State of the art reinforcement learning has enabled training agents on tasks of ever increasing complexity. However, the current paradigm tends to favor training agents from scratch on every new task or on collections of tasks with a view towards generalizing to novel task configurations. The former suffers from poor data efficiency while the latter is difficult when test tasks are out-of-distribution. Agents that can effectively transfer their knowledge about the world pose a potential solution to these issues. In this paper, we investigate transfer learning in the context of model-based agents.
Specifically, we aim to understand \emph{when} exactly environment models have an advantage and \emph{why}. We find that a model-based approach outperforms controlled model-free baselines for transfer learning. Through ablations, we show that both the policy and dynamics model learnt through exploration matter for successful transfer.
We demonstrate our results across three domains which vary in their requirements for transfer: in-distribution procedural (Crafter), in-distribution identical (RoboDesk), and out-of-distribution (Meta-World). Our results show that intrinsic exploration combined with environment models present a viable direction towards agents that are self-supervised and able to generalize to novel reward functions.
\end{abstract}

\section{Introduction}
\label{introduction}

\begin{figure*}[t]
 \centering
    \includegraphics[width=0.95\textwidth]{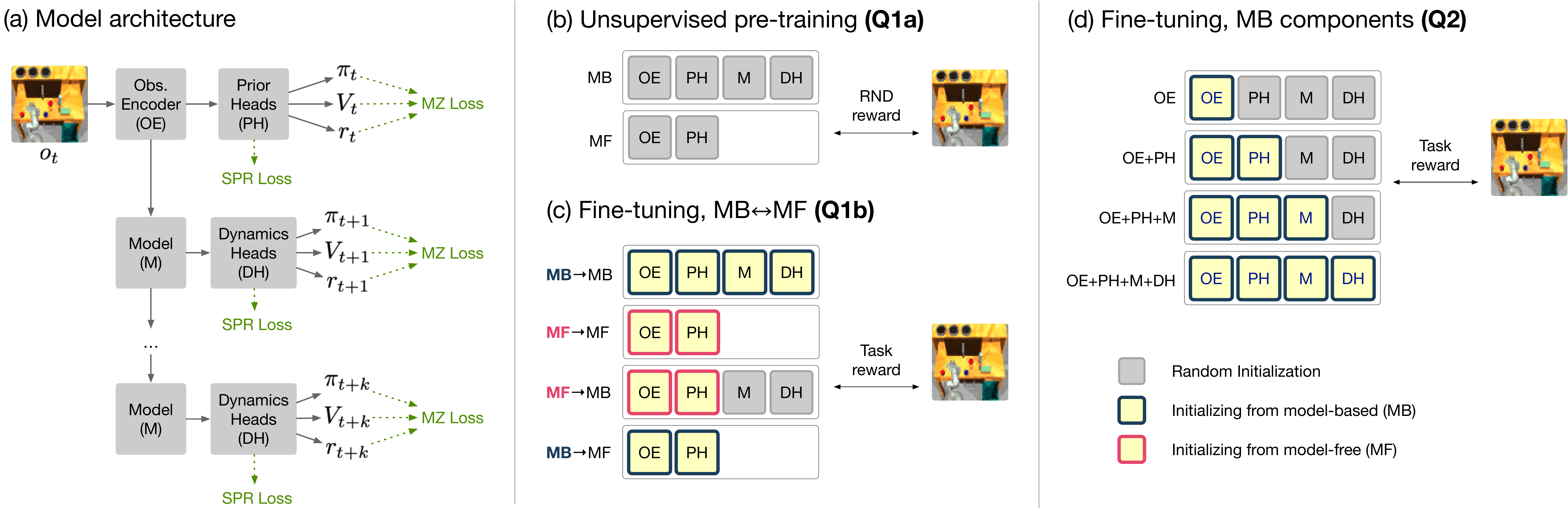}
    \caption{Model-based unsupervised pre-training and fine-tuning. (a) A model-based agent is comprised of an observation encoder (OE), prior heads (PH), model (M), and dynamics heads (DH). These components are trained via a MuZero-style loss \citep{schrittwieser2020mastering} and self-supervision \citep{schwarzer2020data}. (b) During pre-training, the agent interacts with an environment attempting to optimize an intrinsic exploration objective. (c-d) Transfer is evaluated by fine-tuning an agent on a task-based reward in the same environment while carrying over various components from the pre-trained agents. Colored blocks indicate components that are transferred while gray blocks indicate those that are re-initialized. Red indicates transfer from a pre-trained model-free agent, while blue indicates transfer from a pre-trained model-based agent.}
    \label{fig:teaser}
\end{figure*}

A fundamental component of intelligence is generalization: the ability to transfer knowledge to novel situations and tasks. 
Although the field of reinforcement learning (RL) focused for many years on single-task settings in which generalization is not required \citep{mnih2015human,schulman2017proximal,haarnoja2018soft}, there has been a recent surge of interest in designing agents which can successfully transfer their knowledge after training in both multi-task \citep{kirk2021survey,hospedales2021meta} and unsupervised \citep{hansen2019fast, campos2021beyond} settings.
At the same time, another thread of research has focused on developing ever-more powerful model-based agents \citep{schrittwieser2020mastering,hafner2023mastering} which many expect to be an essential ingredient in generalization and transfer \citep{tolman1948cognitive,dayan1995helmholtz,ha2018recurrent,schmidhuber1991curious,sutton1991dyna}.
Indeed, previous work has shown that model-based RL is advantageous for generalization under certain assumptions \citep{sekar2020planning,anand2021procedural}, lending weight to the hypothesis that model-based reasoning is important for transfer.

Although previous work has demonstrated the efficacy of model-based learning for generalization in some cases, it has not shed much insight as to \emph{when} we ought to expect it to help or \emph{why}.
In this paper, we address these questions by studying the performance of agents trained with intrinsic exploration and fine-tuned on downstream tasks (see \autoref{fig:teaser} for an overview of our setup). 
By utilizing unsupervised pre-training via intrinsic exploration, we control the type of knowledge the agent can potentially transfer: it knows about the world it inhabits, but not about possible tasks that are achievable in this world.
To identify how much different components such as the model or policy contribute to transfer performance, we fine-tune with either pre-trained or reinitialized weights; were we in a zero-shot generalization setting, any such ablations would lead to catastrophic (and thus uninformative) losses in performance).

To conduct our experiments, we make a variety of implementation choices regarding model-based and model-free learning and intrinsic exploration; however, we emphasize that our contribution is less about these particular choices and more about the insights that our experiments bring.
For the model-based agent, we employ a self-supervised variant of MuZero \citep{schrittwieser2020mastering} trained with self-predictive representations (SPR; \citealp{schwarzer2020data}).
This variant of MuZero has previously been shown to perform well at zero-shot generalization \citet{anand2021procedural}, thus making it a sensible candidate to test transfer, too.
We contrast its performance with a model-free Q-learning agent based on the same architecture.
During pre-training, both model-free and model-based agents are trained with random network distillation (RND; \citealp{rnd}), a straightforward and robust approach to exploration.
We evaluate transfer across a number of task suites with different characteristics including procedurally generated environments (Crafter; \citealp{hafner2021benchmarking}), settings where the environment in the pre-training is the same (RoboDesk; \citealp{kannan2021robodesk}), and settings where the fine-tuned environment is partially out-of-distribution (Meta-World; \citealp{yu2020meta}).

Overall, we find that model-based exploration combined with model-based fine-tuning results in better transfer performance than model-free baselines.
More precisely, we show that: 
\begin{enumerate}[1)]
    \item model-based methods perform better exploration than their model-free counterparts in reward-free environments; 
    \item knowledge is transferred most effectively when performing model-based (as opposed to model-free) pre-training and fine-tuning;
    \item system dynamics present in the world model seem to improve transfer performance; 
    \item the model-based advantage is stronger when the dynamics model is trained on the same environment. 
\end{enumerate}

\section{Background}
\label{background}

\subsection{Transfer learning in RL}

Learning complex tasks from scratch can be prohibitively expensive, if not impossible, and many approaches have been attempted to alleviate this difficulty.
One approach is to pre-train an agent on high quality data obtained from humans \citep{vinyals2019grandmaster, baker2022video}; however, this can be costly if data is not readily available.
Another approach is to pre-train an agent on a family of (potentially easier) tasks which are related to the task of interest, and then transfer to the target task either directly, within a few attempts, or through fine-tuning \citep{zhu2020transfer,kirk2021survey}.
Generally, the knowledge being transferred can take different forms, such as data (or demonstrations), skills and behaviors, value functions, and dynamics (or world models).
Most recently, a new area of focus has been unsupervised reinforcement learning \citep{watters2019cobra, laskin2021urlb, campos2021beyond} in which an agent is pre-trained using without task rewards using intrinsic exploration and then fine-tuned on a task of interest.
Our work falls into this category and investigates the differences between model-free and model-based unsupervised RL.

A number of previous works have explored directions related to model-based transfer learning, though most do not fit under the umbrella of unsupervised RL.
Some works have investigated training models on source tasks and transferring them to target tasks with MPC, but do not consider policy fine-tuning \citep[e.g.][]{dasari2019robonet,bucher2021adversarial,lutter2021learning,byravan2021evaluating}.
Others do consider policy fine-tuning, but as with most other approaches, utilize a distribution of source tasks rather than unsupervised exploration \citet[e.g.][]{byravan2020imagined}.
Another approach is to use a pre-trained model as a way to speed up policy transfer by learning a policy from interaction with the model \citep[e.g][]{nagabandi2018neural,sekar2020planning}.
Of note is work by \citet{sekar2020planning}, which also incorporates an unsupervised exploration phase for model learning.
However, their approach to transfer relies on being able to relabel previously collected experience with the target reward function, whereas we make no such assumption.

\subsection{Intrinsic exploration}

The goal of intrinsic exploration is to learn a policy which explores the environment in a general manner without the need for an explicit task-based reward function. This permits the agent to learn from environments where reward labels are unavailable or sparse. 
Count-based exploration methods \citep{bellemare2016unifying,ostrovski2017count,tang2017exploration} keep track of how often each state has been visited, and reward the agent for visiting novel or infrequently visited states.
Another approach based on state frequency is to directly maximize the entropy of the state distribution \citep{hazan2019provably,yarats2021reinforcement}.
However, most real-world environments are not tabular and it is therefore not straightforward how to keep track of visitation counts.
As an alternative, curiosity-driven exploration computes a measure of surprise (such as prediction error or uncertainty) and rewards the agent for visiting surprising states \citep{schmidhuber1991possibility,oudeyer2007intrinsic,DBLP:journals/corr/PathakAED17,rnd}.
Ensemble-based methods are a subclass of curiosity-based approaches and estimate uncertainty in a principled way by leveraging an ensemble of predictions \citep{osband2016deep,lowrey2018plan,pathak2019self,sekar2020planning}; however, they are often computationally expensive and challenging to implement.
Finally, instead of seeking surprising states (as in curiosity-based approaches), empowerment focuses on learning skills to control the environment and which support exploring the state space more efficiently \citep{Klyubin2005EmpowermentAU,gregor2016variational,eysenbach2018diversity}.
In this paper, we leverage Random Network Distillation \citep{rnd}, which is a curiosity-based approach in which prediction error is computed using a fixed randomly initialized neural network as the target.
RND is straightforward to implement and has been shown to work robustly across a variety of domains \citep{rnd,laskin2021urlb}.

% \theo{Add a very general blurb about unsupervised RL and different ways to achieve it: curiosity, and empowerment \citep{groth2021curiosity};\citep{eysenbach2018diversity}; \citep{gregor2016variational}, maybe also \citep{balloch2022role}}

\section{Methods}
\label{sec:method}

To investigate the efficacy of model-based transfer, we begin by considering a model-based architecture (\autoref{fig:teaser}a), described in more detail in \autoref{sec:agent}.
During unsupervised \textbf{pre-training} (\autoref{fig:teaser}b), we train agents based on this architecture to explore using an intrinsic motivation signal derived from curiosity, as described in \autoref{sec:rnd}. Then, during \textbf{fine-tuning} (\autoref{fig:teaser}c-d), we transfer various components of these agents and train them on downstream tasks using only the task reward and no intrinsic reward.
Using this framework, we consider the following questions:

\textbf{Q1}: \textit{Is there an advantage to an agent being model-based during unsupervised exploration and/or fine-tuning?}
We investigate this question by first pre-training both model-free and model-based agents to maximize an intrinsic reward in an unsupervised exploration phase. We then fine-tune these agents using either model-based (MB) or model-free (MF) learning, resulting in four variations: MB$\rightarrow$MB, MB$\rightarrow$MF, MF$\rightarrow$MB, and MF$\rightarrow$MF.
We also consider a further baseline, `Scratch', in which the model-based agent is trained on the test task with randomly initialize weight (i.e.\ without any pre-training). Additional details are provided in \autoref{app:agent_details}.
By comparing these different agents, we can determine whether model-based pre-training leads to improved transfer performance.

\textbf{Q2}: \textit{What are the contributions of each component of a model-based agent for downstream task learning?}
We investigate this by looking at various ablations of the full model-based agent (MB$\rightarrow$MB), where only certain components are transferred whereas others are reset.
In particular, we investigate the effect of the dynamics heads (DH), the dynamics model (M), and the prior heads (PH).
We also specifically ablate the prior policy head (PP) apart from the prior reward and value heads (PRV).
By performing these ablations over the various combinations of agent components, we can understand the contributions of each model to downstream task performance.
    
\textbf{Q3}: \textit{How well does the model-based agent deal with environmental shift between the unsupervised and downstream phases?}
Finally, by investigating transfer performance in different classes of environments we can consider the effects of environment mismatch on downstream performance.  We propose three environments to look at: a procedurally generated environment with the same distribution in both the unsupervised and the downstream phase (Crafter; \citealp{hafner2021benchmarking}), one with the same MDP (reward function put aside) during both the unsupervised phase and the downstream phase (RoboDesk; \citealp{kannan2021robodesk}), and one with a stronger shift where system dynamics are maintained but the objects the agent interacts in downstream tasks have various degrees of novelty relative to ones seen during unsupervised exploration (Meta-World; \citealp{yu2020meta}).

\subsection{Agents}
\label{sec:agent}

\paragraph{Model-Based} We use MuZero \citep{schrittwieser2020mastering} as the ``backbone" of our agent.
As MuZero has shown superior generalization capabilities when combined with representation learning techniques, we use a variant of MuZero with self-predictive (SPR) loss as the auxiliary loss \citep{anand2021procedural}.

MuZero learns a partial world model that predicts rewards, actions, and values.
Similar to AlphaZero~\cite{silver2018general}, it then incorporates this model in Monte-Carlo Tree Search~\citep{kocsis2006bandit,coulom2006efficient} to plan and choose optimal actions. The reward, value, and policy are learned as the agent collects experience from the world and plans.
The loss at time-step $t$ is the following:
\vspace{-0.5em}
\begin{align}
l_{t}(\theta) = \sum_{k=0}^K\ & l_{\pi}^k + l_{v}^k + l_{r}^k + l^k_{\text{SPR}} \label{eq:muzero} \\
= \sum_{k=0}^K\ & \mathrm{CE}(\hat{\pi}^k, \pi^{k}) \nonumber
+ \mathrm{CE}(\hat{v}^k, v^{k}) \\
& + \mathrm{CE}(\hat{r}^k, r^{k}) \nonumber
+ \mathrm{CS}(\hat{y}^k, y^{k}),
\end{align}
where $K$ represents the number of steps MuZero plans in the future.
$\hat{\pi}^{k}, \hat{v}^{k}$ and $\hat{r}^{k}$ are predictions for
policy $\pi^{k}$ from the search tree, $v^{k}$ is an $n$-step return bootstrapped 
by a target network, and ${r}^{k}$ is the true reward from the environment. As in the original MuZero~\cite{muzero}, we define the loss for training the value and reward functions via a cross-entropy (CE) distributional RL loss. The SPR loss is computed as a cosine similarity (CS) between the projections predicted by the dynamics model $\hat{y}_k$ and projections computed using observations at timestep $t+k$, $y_k$.
We also use replay via Reanalyse~\cite{reanalyse} for better data efficiency. For environments with a continuous action space, we use sampled MuZero~\cite{sampledmuzero} which plans over sampled actions.
More details of the MuZero agent are provided in \autoref{app:agent_model_based}.

\paragraph{Model-Free} As a model-free baseline for pre-training, we use the Q-Learning agent as described in \citet{anand2021procedural}. The Q-Learning baseline is based on the same architecture as MuZero agent, and in particular, utilizes the same architecture for the observation encoder and prior heads.
This allows us to transfer the component weights from the model-free baseline to the model-based agent, and vice versa (see \autoref{fig:teaser}).
To transfer MuZero weights to a Q-Learning agent (MB$\rightarrow$MF), we transfer only the observation encoder and prior heads.
To transfer Q-Learning weights to a MuZero agent (MF$\rightarrow$MB), we transfer over the observation encoder and prior heads while initializing the dynamics model and dynamics heads from scratch.
More details of the Q-Learning agent are provided in \autoref{app:agent_model_free}.

\subsection{Exploration}
\label{sec:rnd}
To drive exploration during the unsupervised pre-training phase, we use intrinsic rewards computed from Random Network Distillation (RND; \citealp{rnd}). RND can be seen as a method for approximately quantifying uncertainty, and which has proven empirically to be a robust approach to intrinsic exploration \citep{burda2018large,laskin2021urlb}. We found that RND worked well with relatively less tuning with our framework versus approaches such as BYOL-explore \citep{guo2022byol}.

RND defines an intrinsic reward by randomly projecting an observation to feature space $z=f_{\text{rand}}(o)$.
For a given observation $o$, the agent attempts to predict the random projection via $\hat{z}=f_{\theta}(o)$, where $\theta$ are learnable parameters.
The error between the agent's prediction and the random projection, $e = ({z} - \hat{z})^2$, provides an intrinsic reward signal.
Intuitively, since $f_{\text{rand}}$ is not known, the value of $f_{\text{rand}}$ in a state can only be known  (predictable) to the agent if the agent has visited the state before; novel states in the environment so far unvisited by the agent should have large error, and correspondingly high curiosity signal.

To incorporate RND into MuZero, we replace ${r}^{k}$ with ${e}^{k} = ({z}^{k} - \hat{z}^{k})^2$ in \autoref{eq:muzero}, where ${z}^{k}$ is the random projection of observation ${o}^{k}$, and $\hat{z}^{k}$ is the output of the observation encoder given observation $o^k$. Through experimentation we found that a few modifications from the standard RND improved the performance of RND specifically for our architectural setup. We made ${z}$ convolutional features, and our projector for ${z}$ shares the same architecture as the observational encoder for simplicity.

We also investigated using the output of the dynamics model to construct the prediction $\hat{z}^{k}$ which led to inferior exploration performance. 
Note that in contrast to model-free methods, MuZero optimizes the predicted RND signal as opposed to the measured RND signal.

\begin{figure*}
\centering
\includegraphics[width=0.95\textwidth]{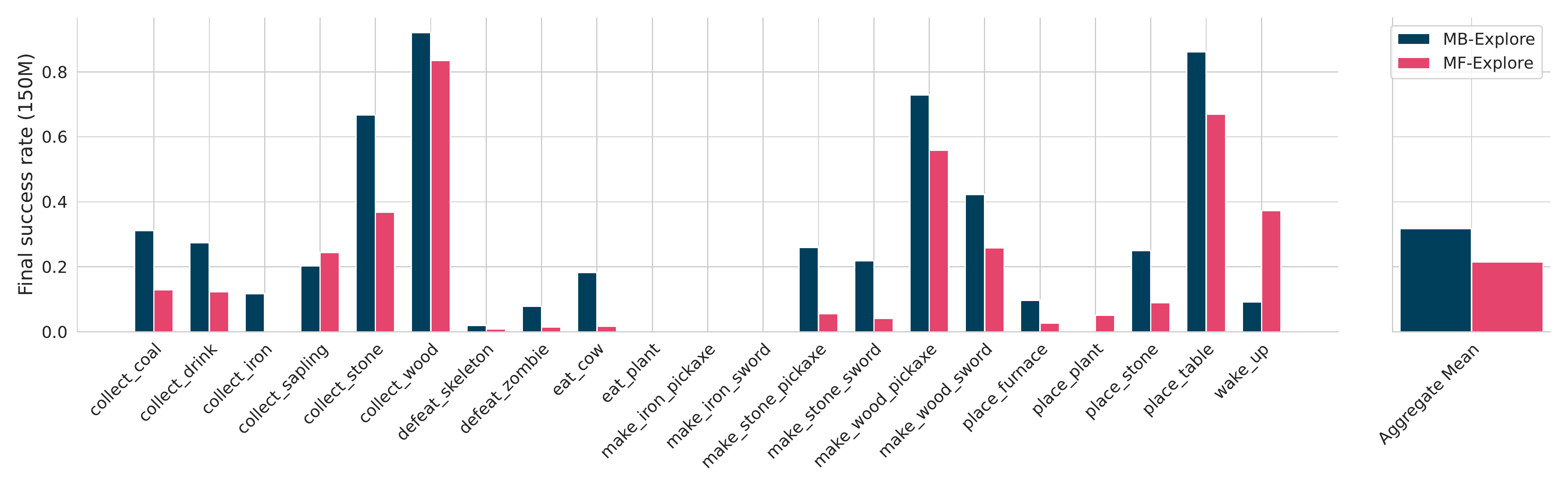}
\caption{Exploration performance of model-based and model-free agents on Crafter as measured by the per-task and aggregate success rates after pre-training for 150 million environment steps.}
\label{fig:crafter_explore}
\end{figure*}

\begin{figure}[t!]
 \centering

 \subfloat[Finetuning return curves]{
 \centering
 \begin{tabular}{ l }
 \includegraphics[width=0.4\textwidth]{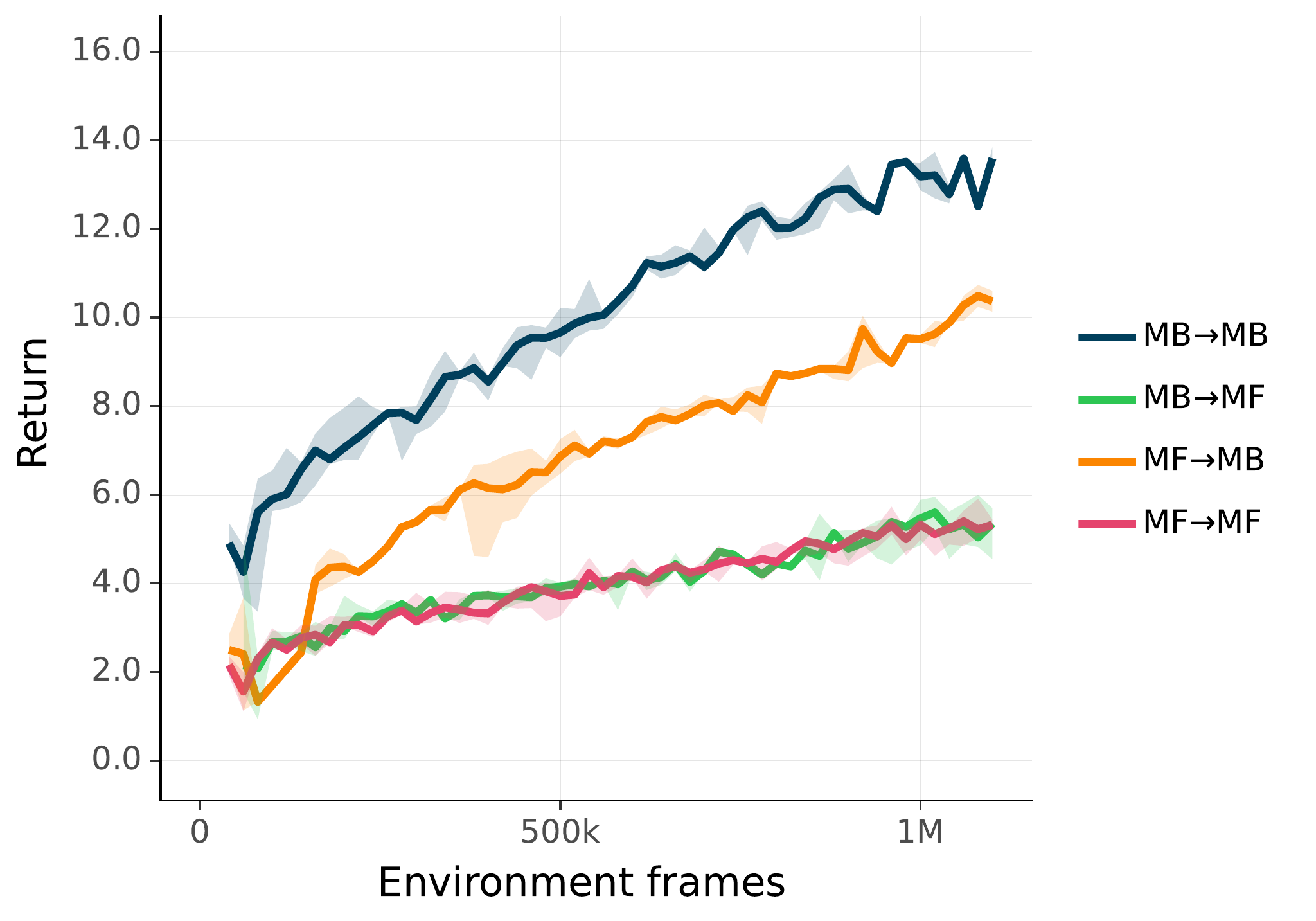} \\ 
 \includegraphics[width=0.43\textwidth]{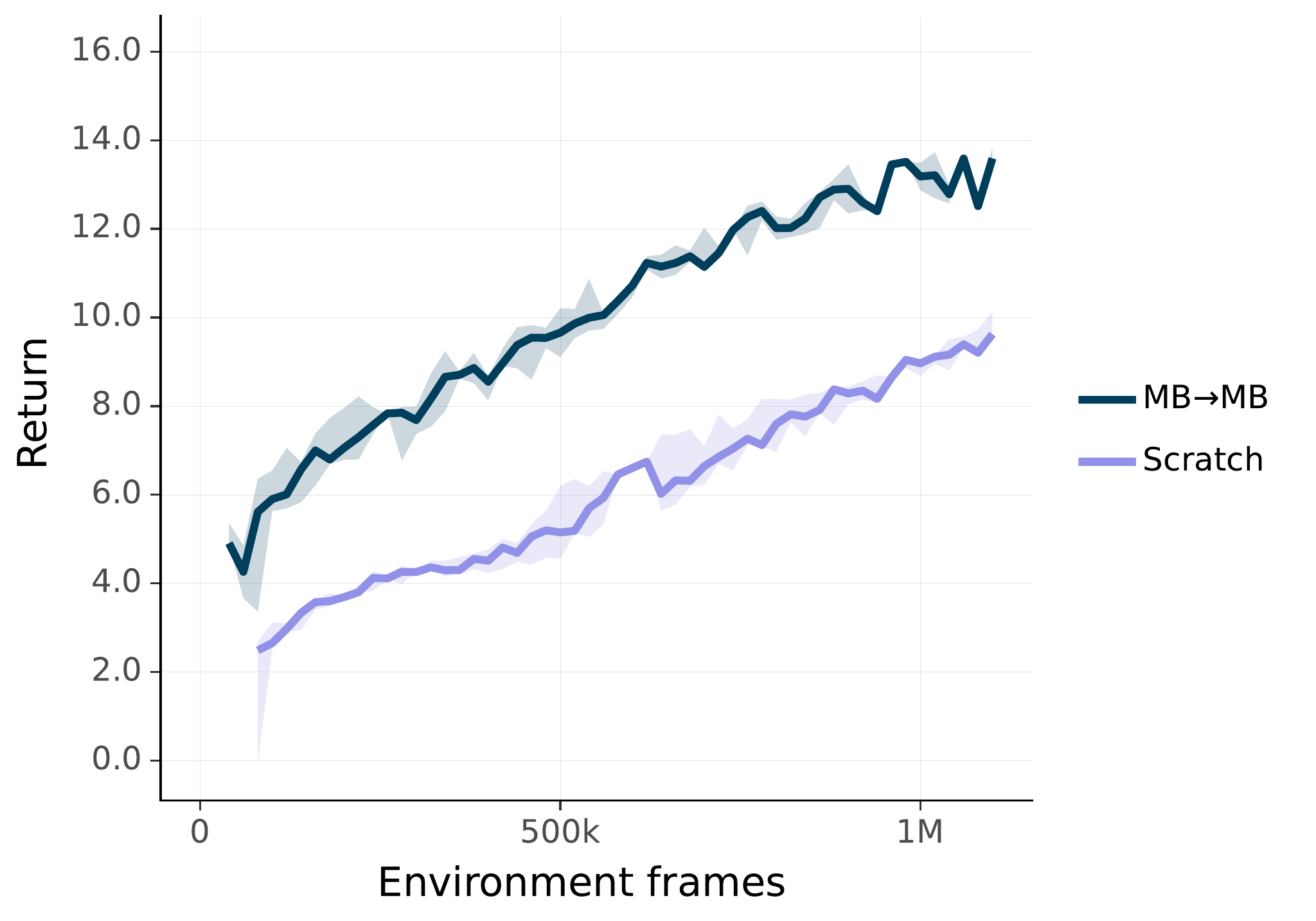} \\ 
 \end{tabular}
 \label{fig:crafter_return}
 }\hfill
  \subfloat[Crafter scores and final rewards]{
  \centering
  \normalsize
 \begin{tabular}{lcc}
 \toprule
 \textbf{Method} & \textbf{Score} & \textbf{Reward}\\
 \midrule
 Human Experts \tiny{\cite{hafner2021benchmarking}} & $50.5\pm6.8$ & $14.3\pm2.3$ \\
 \midrule
 MB$\rightarrow$MB & \textbf{16.4 $\pm$ 1.5} &  \textbf{12.7 $\pm$ 0.4} \\
 MB$\rightarrow$MF & $8.8\pm0.4$ & $5.0\pm0.2$\\
 MF$\rightarrow$MB & $6.2\pm0.5$ & $9.3\pm0.3$ \\
 MF$\rightarrow$MF &  $6.7\pm0.6$ & $6.9\pm0.2$\\
 \midrule
 DreamerV3 \tiny{\cite{hafner2023mastering}} & $14.5\pm1.6$ & $11.7\pm1.9$ \\
 LSTM-SPCNN \tiny{\cite{stanic2022learning}} & $12.1\pm0.8$ & - \\
 DreamerV2 \tiny{\cite{hafner2021benchmarking}} & $10.0\pm1.2$ & $9.0\pm1.7$  \\
 MB Scratch & $4.4\pm0.4$  & $8.5\pm0.1$ \\
 \bottomrule
 \end{tabular}
 \label{fig:crafter_score}
 }
  \caption{
  Agent performance on Crafter. \subref{fig:crafter_return} Return as a function of the environment steps; \subref{fig:crafter_score} Comparison of score and final reward at 1M steps across different fine-tuned agents and published results for agents without fine-tuning.
  }
  \label{fig:finetuning}
  \vspace{-1em}
\end{figure}

\begin{figure}[t!]
\centering
\subfloat[Average success rate]{
\centering
 \includegraphics[width=0.7\linewidth]{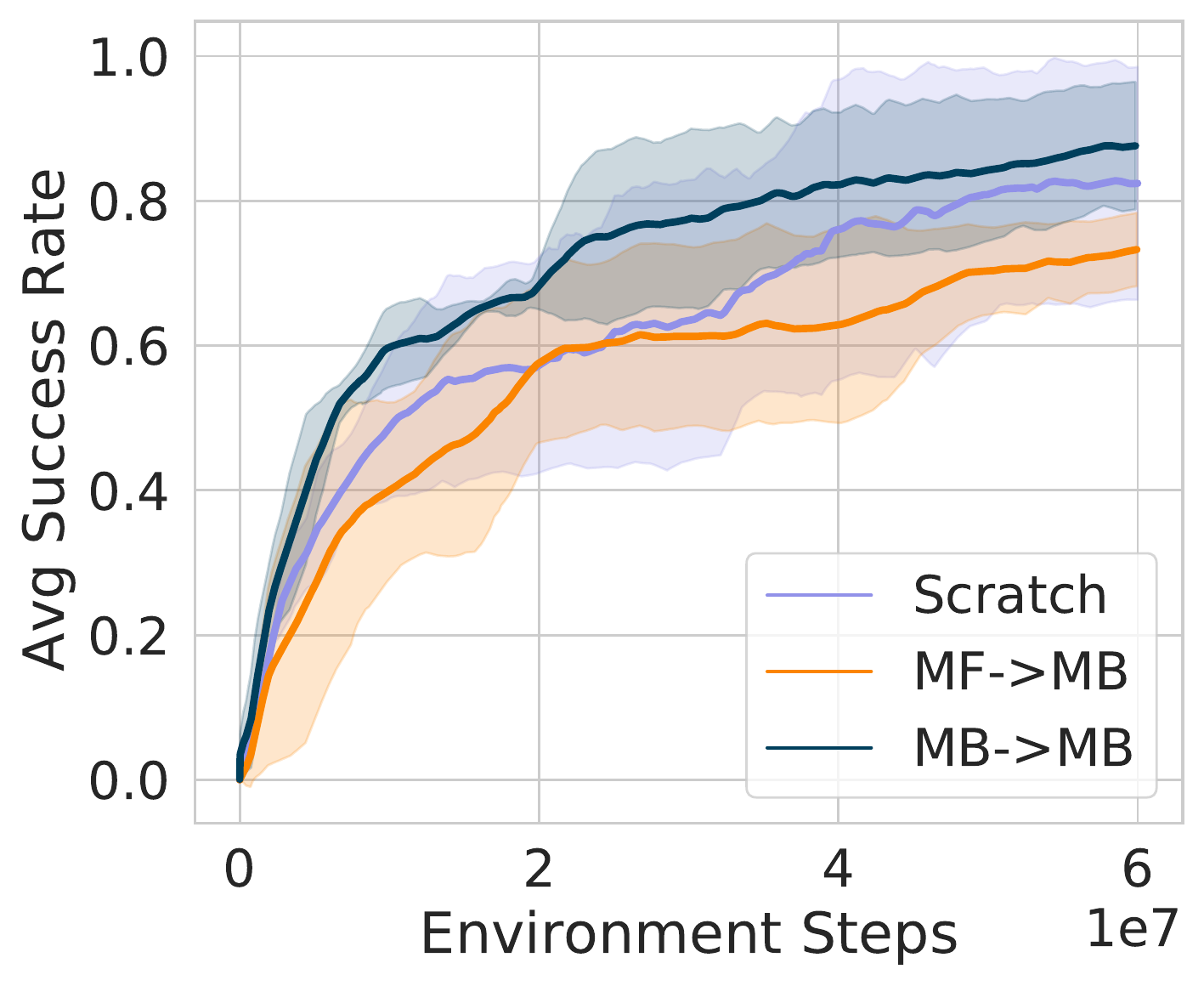}
  \label{fig:robodesk_avg}
  \quad}\\
\subfloat[Success rate on 9 individual tasks]{
 \centering
 \includegraphics[width=\linewidth]{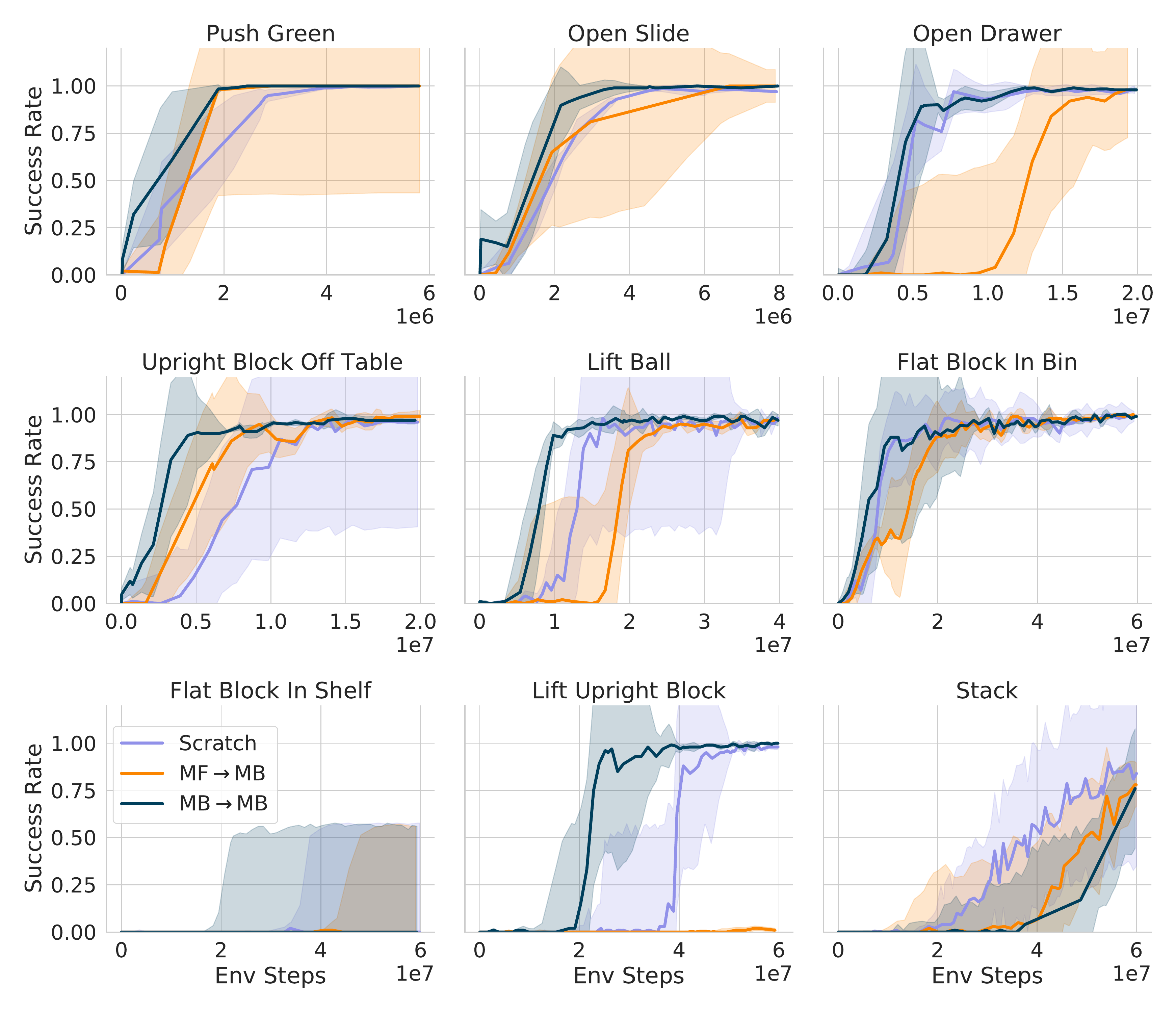}
  \label{fig:robodesk_individual}}
  \caption{Success rate on RoboDesk. \subref{fig:robodesk_avg} Average success rate over 9 tasks and 3 seeds; \subref{fig:robodesk_individual} Success rate on individual tasks by taking the median over 3 seeds. We also report the standard deviation in both plots. Scratch: MuZero trained from scratch on individual tasks with extrinsic reward; MF $\rightarrow$ MB: model-free exploration transferred to MuZero agent fine-tuned on individual task; MB $\rightarrow$ MB: model-based exploration transferred to MuZero agent fine-tuned on individual task.
  }
  \label{fig:robodesk}
\end{figure}

\subsection{Environments}

We conduct our transfer experiments in three settings. The first, Crafter \citep{hafner2021benchmarking}), has the fine-tuning environments in-distribution with the pre-training phase, but episodes are procedurally generated (so that test environments are never encountered during training); the second RoboDesk, \citep{kannan2021robodesk}), is the same environment during pre-training and fine-tuning phase, the only difference being the availability of tasks rewards during the fine-tuning phase; the third, Meta-World, \citep{yu2020meta}) has fine-tuning environments which share some similarities with the pre-training environment, but are nevertheless different (and therefore out of distribution).

\textbf{Crafter} \citep{hafner2021benchmarking} is a survival game inspired by the popular game Minecraft. In Crafter, the agent inhabits a two-dimensional procedurally generated world. The environment is multi-task, and these tasks are hierarchical. In order to achieve certain tasks, the agent must complete other tasks first. These include gathering resources, building tools, and defending against potential threats. There are 22 potential achievements. The reward signal comprises $+1$ for each task achieved for the first time in an episode, and, in addition, the agent needs to maintain a given health level to survive by e.g. eating and drinking. As in the original paper, we report the success rate as the fraction of all training episodes up to 1M environment steps where the agent has achieved the task at least once.
As in \citet{hafner2021benchmarking}, we also report the score, which is the geometric mean of the success rates: $S= \exp\left( \frac{1}{N} \sum_{i=1}^N \log(1 + s_i)\right)- 1$.
Unlike the return, the score favors unlocking difficult achievements over repeatedly completing easier ones, and being an aggregate measure, it also reflects data-efficiency of the agent.

Although we constrain the agent to use 1M environment steps for fine-tuning, we do not impose such constraint on the data consumption during unsupervised exploration. Our focus is on a relative analysis of model-based transfer and not necessarily to benchmark the efficacy of exploration techniques. 

\textbf{RoboDesk} \citep{kannan2021robodesk} is a control environment simulating a robotic arm interacting with a table and a fixed set of objects. The benchmark features nine core tasks with consistent dynamics but randomized object locations. During exploration, the agent can learn to interact with the objects via optimising intrinsic rewards. Then, we fine-tune the agent individually on the nine different task-rewards. Note that there are no novel objects introduced in the fine-tuning phase; the only differences between pre-training and fine-tuning are the rewards.
As the action space for Robodesk is continuous, we adapted our model-based agent according to sampled MuZero~\cite{sampledmuzero} for this environment. Similar to the experimental setup of \citet{dadashi2021continuous}, the episode length is 2000 environment steps with action repeat of 5. However, we use the pixel observations only and sparse rewards for our study.

\textbf{Meta-World} \cite{yu2020meta} is a robotic control suite of up to 50 tasks with a SAWYER arm. We focus on a Meta-World v2 benchmark intended for task generalization, ML-10. In this benchmark, the training suite consists of ten tasks that differ from the testing suite of five tasks.
Meta-world provides different challenges for model-based transfer compared to Crafter and RoboDesk:
while the robotic arm is shared among all environments, the test environments include unseen objects and configurations which allows us to repurpose this benchmark to study model-based transfer in an out-of-distribution configuration. Note that unlike in the original ML-10 benchmark, our agents do not receive reward observations in the training environments.

We use pixels from the \texttt{corner3} angle as observations and do not use state information from the robot arm. We use sparse rewards which are the average episodic task success rates. We did not find any substantial differences in performance between approaches when using dense reward as these likely made the tasks relatively easy to solve. We explore and fine-tune on about 85 million frames. We also use sampled MuZero~\cite{sampledmuzero} in this environment.

\begin{figure}[t!]
 \centering
 \includegraphics[width=0.47\textwidth]{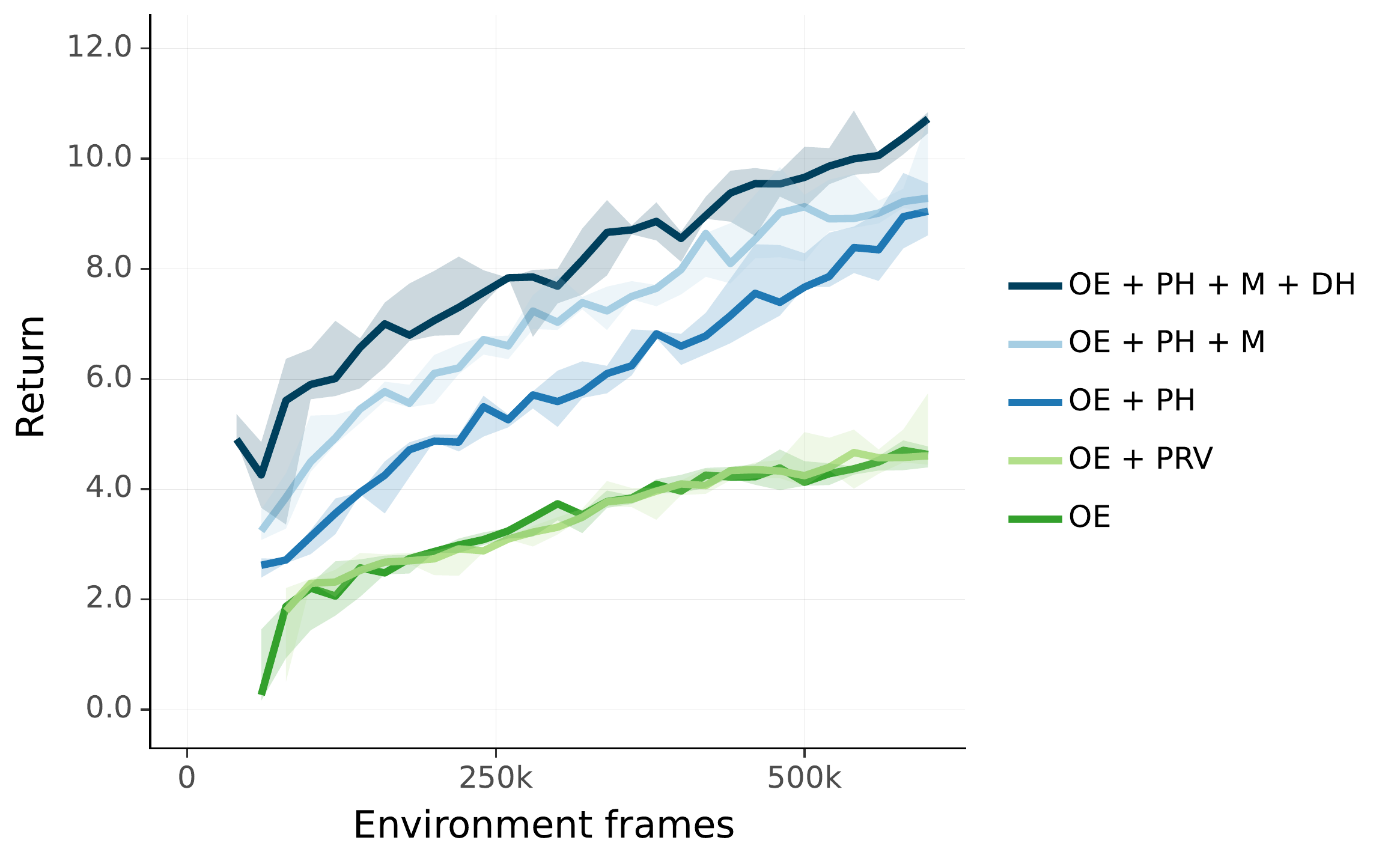}%
  \caption{Here we perform an ablation analysis of which transferred components contribute to improved performance on Crafter. OE represents transferring only the observation encoder, OE + PRV additionally adds the prior reward and value heads, OE + PH adds prior reward, value, and policy heads, OE + PH + M adds the dynamics function, and OE + PH + M + DH adds all (equivalent to MB$\rightarrow$MB). For ablations, we use finetuning parameters specified in the appendix in ~\autoref{tab:hyperparams_mb}.
  }
  \label{fig:ablation}
  \vspace{-1em}
\end{figure}

\section{Results}
\label{sec:experiment}

Through our experiments, we generally find that model-based exploration and fine-tuning outperforms model-free approaches.
In what follows, we investigate the questions from \autoref{sec:method} in detail.
We restate each question, give a summary of our answer, and then explain the results in more depth.

\textbf{Q1}: \textit{Is there an advantage to an agent being model-based during unsupervised exploration and/or fine-tuning?}

\textbf{A1}: \textit{Yes. Compared to model-free exploration, model-based exploration is more performant and transfers more effectively. Model-based fine-tuning also outperforms model-free fine-tuning.}

To answer this question, we compare different combinations of model-based and model-free pre-training and fine-tuning (i.e., MB$\rightarrow$MB, MB$\rightarrow$MF, MF$\rightarrow$MB, MF$\rightarrow$MF, and Scratch described in \autoref{sec:method}).

In Crafter (\autoref{fig:finetuning}), we find that model-based pre-training with model-based fine-tuning (MB$\rightarrow$MB) substantially outperforms all other agent variations including training from scratch, achieving a score of $16.4\pm 1.5$ and reward of $12.7\pm 0.4$ at 1M environment steps.
In fact, this result improves performance over state-of-the-art model-based agents without a pre-training phase, namely DreamerV3 \citep{hafner2023mastering}, and demonstrates the benefits of combining model-based learning with unsupervised pre-training.
We find that the fully model-free agent (MF$\rightarrow$MF) performs worst, consistent with other results demonstrating inferior model-free performance in other generalization settings \citep{anand2021procedural}.
However, these results raise the question: is the improved performance of MB$\rightarrow$MB due to pre-training, fine-tuning, or both?
The performance of the MF$\rightarrow$MB agent suggests that a good enough agent can leverage knowledge acquired during pre-training, even if the pre-training was with an inferior algorithm.
However, the performance of the MB$\rightarrow$MF agent---which provides no improvement over MF$\rightarrow$MF---suggests that the fine-tuning algorithm plays an important role, too.
It also suggests that additional knowledge may be encoded in the dynamics components, which the model-free agent cannot use.

As Crafter is, in part, designed to evaluate agents' ability to explore efficiently in the absence of reward, we have also quantified differences between the model-based and model-free exploration agents during pre-training. While both agents aim to maximise RND intrinsic rewards, model-based optimisation leads to significantly more effective exploration as measured by the average  and individual success rates across achievements (see \autoref{fig:crafter_explore}).

In RoboDesk, we fine-tuned the pre-trained agents separately on nine core tasks and see a similar pattern of results as in Crafter.
Due to resource limitations, we focused our comparisons on the agent trained from scratch and the two agents that provided the best transfer in Crafter, namely MB$\rightarrow$MB and MF$\rightarrow$MB.
We find that fine-tuning from model-based exploration (MB$\rightarrow$MB) consistently outperforms fine-tuning from the model-free counterpart (MF$\rightarrow$MB) (\autoref{fig:robodesk}).
Transferring from model-free also appears to result in a larger variance during fine-tuning.
This, again, suggests model-based pre-training is superior to model-free pre-training. 
Additionally, compared to MuZero training from scratch (\autoref{fig:robodesk_avg}), warm-starting from the pre-trained weights of the model-based exploration agent shows benefits in terms of sample-efficiency but not in final performance.
This is more evident on tasks such as \textit{upright block off table}, \textit{lift ball}, \textit{lift upright block}, and \textit{flat block in shelf} (\autoref{fig:robodesk_individual}).
Interestingly, these are relatively difficult tasks to solve because they involve multiple steps or more challenging object manipulation.
The improvements are less obvious on easy tasks (e.g. \textit{push green}, \textit{open slide}, \textit{open drawer}).
Similarly to Crafter, we found that model-based exploration outperformed its model-free counterpart when evaluating success-rates across different tasks during pre-training (\autoref{app:result_robodesk}). When looking across tasks, we did not observe a correlation between success rates during pre-training and positive transfer during fine-tuning, suggesting that transfer is not mediated by the alignment of intrinsic and task rewards.

\textbf{Q2}: \textit{What are the contributions of each component of a model-based agent for downstream task learning?}

\textbf{A2}: \textit{Both the dynamics components (model and dynamics heads) and the prior heads (in particular, the prior policy) play an important role in transfer performance.}

To answer this question, we return to the Crafter environment and perform an ablation analysis on MB$\rightarrow$MB where different parts of the fine-tuning agent are initialized from the model-based exploration agent.
These ablations involve the observation encoder (OE), the prior heads (PH), the model (M) and the dynamics heads (DH), as described in \autoref{fig:teaser} and \autoref{sec:method}.
In \autoref{fig:ablation}, we show the performance of the fine-tuning agent when incrementally removing these various elements.
First, removing the dynamics heads from the full agent results in a small drop in performance.
Additionally removing the dynamics model results in a further drop, indicating that the dynamics components (both DH and M) do encode useful knowledge that is leveraged during transfer.
However, it is not the \emph{only} knowledge that can be transferred.
When we remove the prior heads, we find performance further deteriorates.
Moreover, this seems to be driven mostly by the policy prior (PP), as performance appears to be the same regardless of whether we include the prior reward and value heads (PRV) or not.
In our model-based agent based off of MuZero, the policy prior plays an important role in guiding action selection in the search tree and, consistent with that, these ablations show that transferring both the dynamics model and policy prior contribute most to positive transfer.
This result suggests that other model-based approaches to transfer which only transfer the model \citep[e.g.][]{sekar2020planning} could benefit from transferring policy components as well.

\textbf{Q3}: \textit{How well does the model-based agent deal with environmental shift between the unsupervised and downstream phases?}

\textbf{A3}: \textit{Model-based agents can successfully transfer knowledge to novel tasks when the dynamics are identical or in-distribution to those observed during training; however, transfer works less well to out-of-distribution dynamics.}

\begin{figure}[t!]
 \centering
 \begin{tabular}{ l }
 \includegraphics[width=0.4\textwidth]{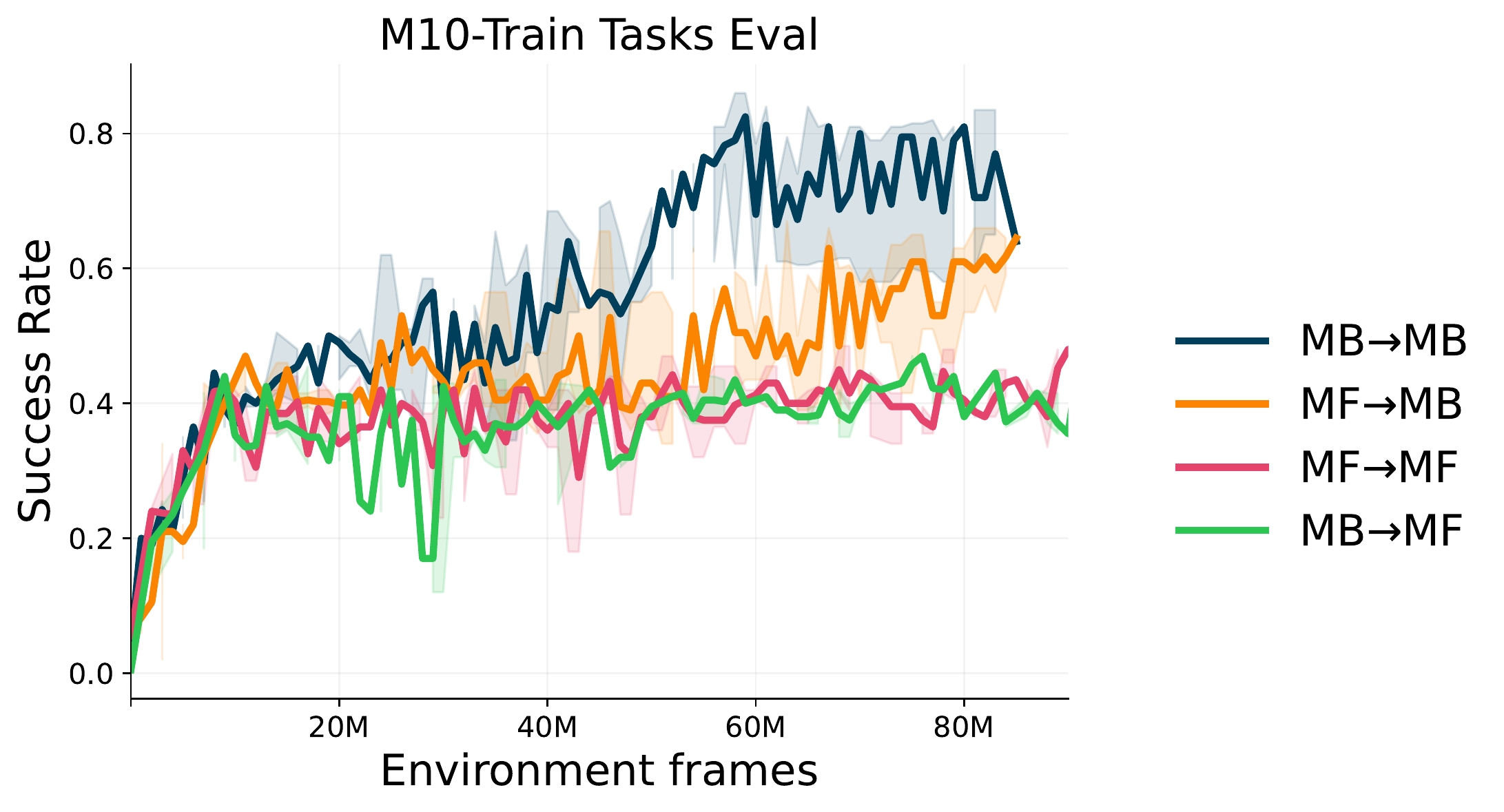} \\ 
 \includegraphics[width=0.4\textwidth]{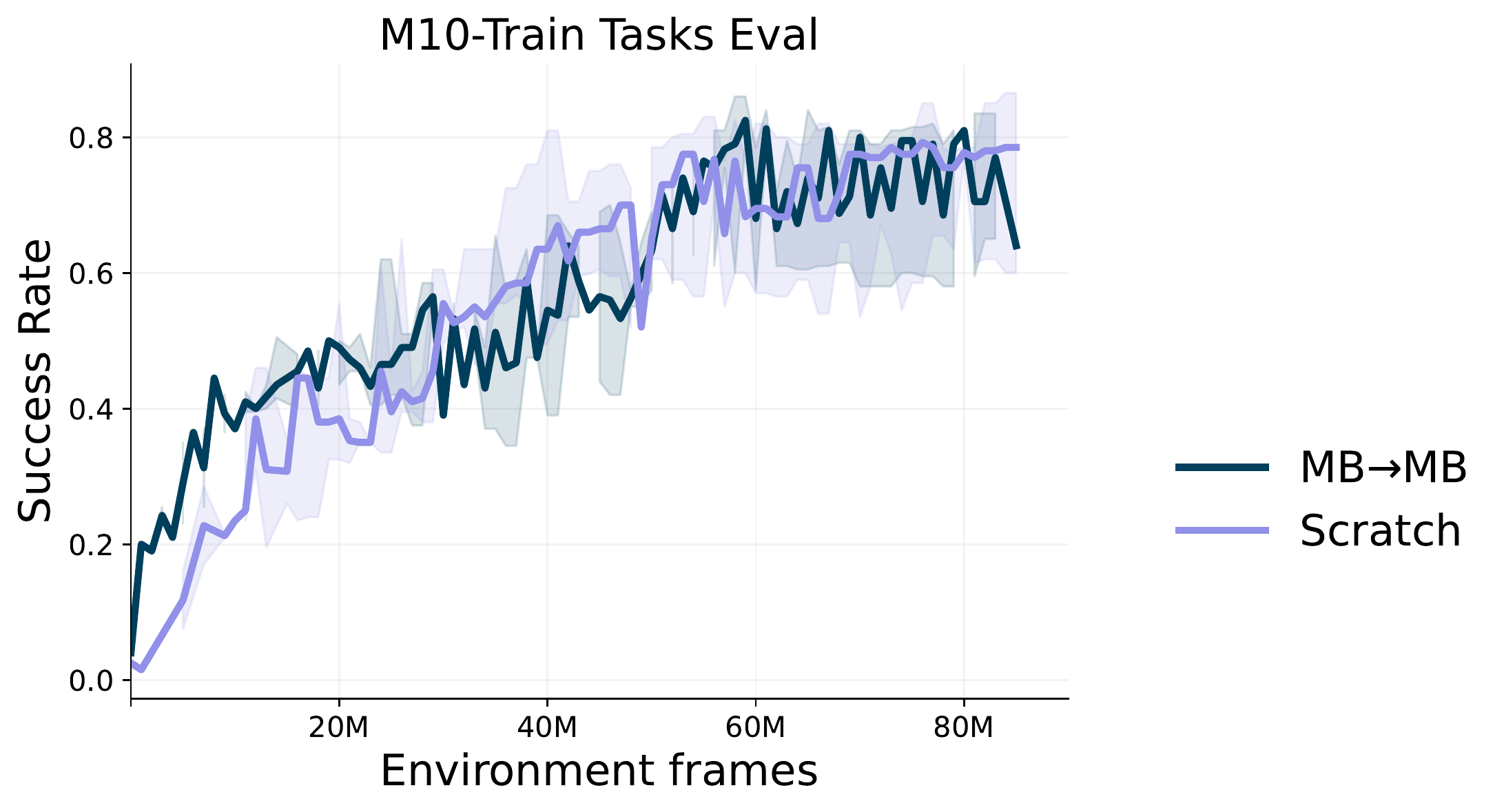} \\ 
 \end{tabular}
  \caption{Here we report the return curves (top) of various fine-tuning strategies on Meta-World ML-10 test set.}
  \label{fig:finetuning_metaworld}
  \vspace{-1em}
\end{figure}

Our results on Crafter and, to some extent, Robodesk suggest that model-based agents can successfully transfer knowledge from an unsupervised exploration phase to new reward functions when the pre-training and fine-tuning environments are identical or come from the same distribution. This setting is the most likely to benefit from learning and transferring dynamics models as the train and test environments differ only in the reward function.
To better understand the limits of model-based transfer in the case of out-of-distribution environments, we also run a set of transfer experiments on the ML-10 Meta-World benchmark.
Similarly to our results on Crafter and Robodesk, we find that that MB$\rightarrow$MB outperforms other pre-trained baselines (\autoref{fig:finetuning_metaworld}, top). However, when compared with an agent trained from random initialization, MB$\rightarrow$MB has a small advantage only early in training, up to $\sim$20 million frames (\autoref{fig:finetuning_metaworld}, bottom). Interestingly, model-free baselines appear to have inhibited transfer performance in this domain vs randomly initialized, and model-based transfer only helps mildly. We speculate this difference in results between Meta-World and the other two benchmarks is due to out-of-distribution environment dynamics. Due to the different objects between the train and test environment the agent is able to transfer its knowledge only about the dynamics in the robotic arm, and our empirical results suggest that that alone may be insufficient for positive transfer.  

\section{Discussion}
In this paper we propose to study \emph{when} and \emph{why} model-based learning is beneficial for transferring knowledge. We make use of a framework which consists of an unsupervised pre-training phase in a no reward environment, followed by fine-tuning with task-specific reward functions.   We argue that this is a suitable framework to ask (and answer) these questions. We present a specific instantiation of a model-based agent as a strong baseline, as well as a model-free counterpart using the same ``backbone''. 
We study a number of key factors that may contribute to successful transfer, namely the agent's ability to \emph{explore} the environment even in the absence of any rewards and to summarize its experience and knowledge about the environment in the form of \textit{representations}, \emph{policies} and \emph{dynamics models} that lend themselves to transfer. 
By conducting the same experiments on three distinct environments (in-distribution
procedural, in-distribution identical, and out-of-distribution), we investigate transfer under environment shift. 

Overall, we find that model-based optimization dominates model-free variants both during unsupervised exploration and fine-tuning performance. Our analysis reveals that transferring the dynamics model (and heads) as well as the prior policy learned during exploration contributes most to transfer.
We observe that environment shift between unsupervised and downstream phases is detrimental for knowledge transfer. We speculate that the significantly stronger transfer results on Crafter, compared to Robodesk, are enabled partly by the procedural environment variations seen during pre-training.

While the exact design choices are reflected in our quantitative results, we expect that this framework yields more broadly applicable qualitative conclusions as well.
Throughout this paper we have focused on a model-based agent derived from MuZero, however, we would expect further benefits from using a more sample efficient variant like EfficientZero \cite{ye2021mastering} or DreamerV3 \cite{hafner2023mastering} as the base agent.
Our study fixes the intrinsic exploration mechanism, this is because we focus on contrasting model-based versus model-free approaches rather than different choices of the intrinsic reward. Since our investigation shows that policy transfer matters, alternative approaches such as empowerment maybe beneficial.

\section*{Acknowledgements}

We would like to thank Michael Laskin, Loic Matthey, and Simon Osindero for helpful discussions and feedback.

\bibliography{example_paper}
\bibliographystyle{icml2023}

\newpage
\appendix
\onecolumn
\section{Agent Details}
\label{app:agent_details}
\subsection{Network Architecture} 

Both the model-free and model-based agents utilize the same network architectures. It's identical to the ones used in MuZero ReAnalyse \citep{reanalyse}. 
The pixel input resolution is $96 \times 96$ for Crafter and Meta-World; $64 \times 64$ for RoboDesk. The images are first sent through a convolutional stack that downsamples to an $6\times6$ tensor (for Meta-World) or $8\times8$ tensor (for RoboDesk and Crafter). 
This tensor then serves as input to the encoder. Both the encoder and the dynamics model were implemented by a ResNet with 10 blocks, each block containing 2 layers.  Each layer of the residual stack was convolutional with a kernel size of 3x3 and 256 planes. 

\subsection{Model-based Agent}
\label{app:agent_model_based}
\paragraph{MuZero + SPR}

To add the SPR as an auxiliary loss, we add a projection and a prediction head, similar to \citet{grill2020bootstrap}.
The projection and prediction networks have an identical architecture: a two convolutional layers with stride 1 and kernel size 3 and relu non-linearity in between; then flatten the output to a vector.
The input to the projection network is the output of dynamics function; the input to the prediction network is the output of the projector. 
Since the dynamics model unrolls $n$ steps into the future, it results in $n$ prediction vectors $x_{n}$.
The target vector $y_n$ is the output of the projector computed with the target network weights. The target network weights is the same as the one used to compute the temporal difference loss in MuZero.
The input for computing the target vectors are the corresponding image observation at that future step $n$. 
The encoder of MuZero uses 15 past images as history. However, when we compute the target vectors, our treatment of the encoding history is different from that of the agent; instead of stacking all of the historical images ($n-15...n-1$) up to the corresponding step $n$, we simply replace the history stack with 15 copies of the image at the current step $n$. This is the same treatment as in \citet{anand2021procedural}.
We then attempt to match $x_{n}$ with the corresponding target vector $y_{n}$ with a cosine distance between $x_i$ and $y_i$.

\paragraph{Hyper-parameters}

\Cref{tab:hyperparams_mb} listed the hyper-parameters of our model-based agent for both pre-training and fine-tuning. If no special indication in \Cref{tab:hyperparams_mb}, pre-training and fine-tuning use the same value.
For all the agents, we use the Adam \citep{kingma2014adam} optimizer.
The batch size for the training is $1024$. The Scratch baseline uses the pre-train hyper-parameters, but always with initial learning rate of $10^{-4}$ and cosine learning rate schedule. 

\begin{table}[h!]
\caption{Hyper-parameters for Model-based Pre-training and Fine-tuning}
\label{tab:hyperparams_mb}
\begin{center}
\begin{small}
\begin{tabular}{lccc}
\toprule
\midrule
\textsc{Hyper-parameter} & \textsc{Crafter} & \textsc{RoboDesk} & \textsc{Meta-World} \\
& (\textit{Pre-train / Fine-tune}) & (\textit{Pre-train / Fine-tune}) & (\textit{Pre-train / Fine-tune}) \\
\midrule
\textsc{Training} & & \\
\midrule
Model Unroll Length & $5$ & $5$ & $5$ \\
TD-Steps & $5$ & $0$ & $0$ \\
ReAnalyse Fraction & $0.8$ / $0.99$ & $0.925$ & $0.925$\\
Replay Size (in sequences) & $50000$ & $2000$ & $2000$ \\
\midrule
\textsc{MCTS} & & \\
\midrule
Number of Simulations & $50$ & $50$ & $50$ \\
UCB-constant & $1.25$ & $1.25$ & $1.25$ \\
Number of Samples & n/a & $20$ & $20$ \\
\midrule
\textsc{Self-Supervision} & & \\
\midrule
SPR Loss Weight & $1.0$ & $1.0$ & $1.0$ \\
\midrule
\textsc{Optimization} & & \\
\midrule
Initial Learning Rate & $10^{-4}$ / $10^{-5}$ & $10^{-4}$ & $10^{-4}$ / $10^{-5}$  \\
Learning Rate Schedule & cosine / constant & constant / cosine & cosine  \\
\midrule
\bottomrule
\end{tabular}
\end{small}
\end{center}
\end{table}

\subsection{Model-free Agent}
\label{app:agent_model_free}
\paragraph{Q-Learning} The Q-Learning setup is identical to \citet{anand2021procedural}. We describe here for completeness of the paper.
Our controlled Q-Learning agent has an identical network architecture to the model-based agent, but modifies it in a few key ways to make it model-free rather than model-based.

The Q-Learning baseline uses $n$-step targets for action value function. Given a trajectory $\{s_t, a_t, r_t\}_{t=0}^T$, the target action value is computed as follow
\begin{equation}
 Q_{target}(s_t, a_t)=\sum_{i=0}^{n-1} \gamma^i r_{t+i} + \gamma^n \max_{a \in \mathcal{A}} Q_{\xi}(s_{t+n}, a)  \label{eq:n-step_backup}
\end{equation}
Where $Q_{\xi}$ is the target network whose parameter $\xi$ is updated every 100 training steps.

In order to make the model architecture most similar to what is used in the MuZero agent, we decompose the action value function into two parts: a reward prediction $\hat{r}$ and a value prediction $V$, and model these two parts separately. The total loss function is, therefore,  $\mathcal{L}_{total}=\mathcal{L}_{reward}+\mathcal{L}_{value}$. The reward loss  is exactly the same as that of MuZero. For the value loss, we can decompose \autoref{eq:n-step_backup} in the same way:

\begin{align}
    Q_{target}(s_t, a_t) &= \hat{r}_t + \gamma V_{target}(s) \nonumber \\
    & = \sum_{i=0}^{n-1} \gamma^i r_{t+i} + \gamma^n \max_{a \in \mathcal{A}} \left( \hat{r}_{t+n} + \gamma V_{\xi}(s^{'}) \right) \nonumber \\
    \implies V_{target}(s) &=\sum_{i=1}^{n-1} \gamma^{i-1} r_{t+i} + \gamma^{n-1} \max_{a \in \mathcal{A}} \left( \hat{r}_{t+n} + \gamma V_{\xi}(s^{'}) \right) \label{eq:q_learning_target}
\end{align}

Since the reward prediction should be taken care of by $\mathcal{L}_{reward}$ and it usually converges fast, we assume $\hat{r}_t=r_t$ and the target is simplified to \autoref{eq:q_learning_target}. We can then use this value target to compute the value loss $\mathcal{L}_{value}=\textrm{CE}(V_{target}(s), V(s))$.

In RoboDesk and Meta-World, since it has a continuous action space, maximizing over the entire action space is infeasible. We follow the Sampled Muzero approach \citep{sampledmuzero} and maximize only over the sampled actions.

\paragraph{Hyper-parameters}

\Cref{tab:hyperparams_mf} lists the hyper-parameters for our model-free baselines. For the model-free agent, we only unroll a single-step for computing action and values.

\begin{table}[h!]
\caption{Hyper-parameters for Model-free Pre-training and Fine-tuning}
\label{tab:hyperparams_mf}
\begin{center}
\begin{small}
\begin{tabular}{lccc}
\toprule
\midrule
\textsc{Hyper-parameter} & \textsc{Crafter} & \textsc{RoboDesk} & \textsc{Meta-World} \\
& (\textit{Pre-train / Fine-tune}) & (\textit{Pre-train}) & (\textit{Pre-train / Fine-tune}) \\
\midrule
\textsc{Training} & & \\
\midrule
Model Unroll Length & $1$ & $1$ & $1$ \\
TD-Steps & $5$ & $1$ & $1$ \\
ReAnalyse Fraction & $0.75$ / $0.99$ & $0.945$ & $0.9325$\\
Replay Size (in sequences) & $50000$ & $2000$ & $2000$ \\
\midrule
\textsc{Self-Supervision} & & \\
\midrule
SPR Loss Weight & $1.0$ & $1.0$ & $1.0$ \\
\midrule
\textsc{Optimization} & & \\
\midrule
Initial Learning Rate & $10^{-4}$ / $10^{-5}$ & $10^{-4}$ & $10^{-4}$ / $10^{-5}$  \\
Learning Rate Schedule & cosine / constant & constant / cosine & cosine  \\
\midrule
\bottomrule
\end{tabular}
\end{small}
\end{center}
\end{table}

\subsection{Random Network Distillation}
Our modified version of RND utilizes the same encoder architecture as the agent network described previously. A projector network takes the output of the encoder and projects into a vector. The projector network is of the same architecture as the projector/predictor described for SPR. The target vector $z$ is computed with randomly initialized weights. The prediction vector $\hat{z}$ is computed with the learned weights. The observation encoder receives training signals from both the RND loss and the usual MuZero losses. The prediction error is the L2 distance between the prediction and target $e=(z-\hat{z})^2$.

The intrinsic reward is computed as a function of the prediction error $e$. 
We keep an exponential moving average of the error with a decay of 0.99 and bias correction technique applied \citep{kingma2014adam}. This gives us the mean and standard deviation $\hat{e}$ and $\hat{\sigma}_e$. Then the reward is the normalized prediction error $r_{intrinsic}=(e-\hat{e})/\hat{\sigma_e}$. We do not apply any clipping to the reward.

\section{Additional Results}
\subsection{Crafter}

\begin{figure}[h!]
\centering
\includegraphics[width=1.0\textwidth]{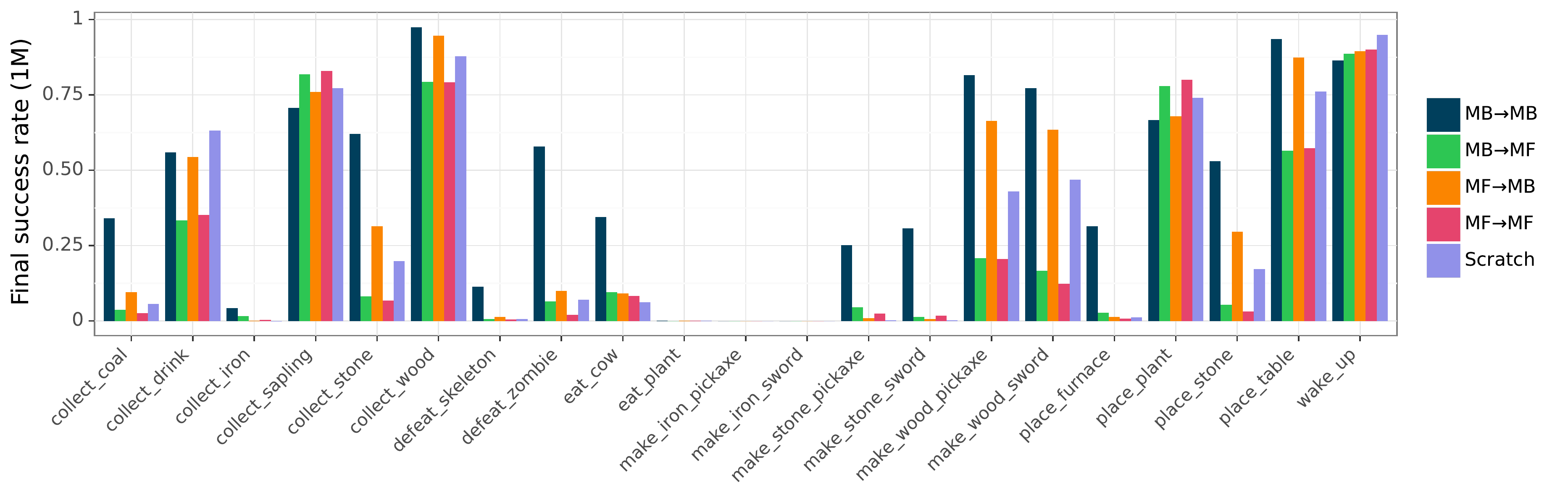}%
 \caption{Success rates on Crafter after fine-tuning. Comparison of sub-task success rates of various transfer agents and baselines.}
 \label{fig:crafter_success}
\end{figure}

\Cref{fig:crafter_success} reports the success rates of the sub-tasks after fine-tuning. We find that the performance margins of MB$\rightarrow$MB are particularly large on the more advanced tasks---especially those requiring or involving stone.

\subsection{RoboDesk}
\label{app:result_robodesk}
\begin{figure}[t!]
\centering
\includegraphics[width=1.0\textwidth]{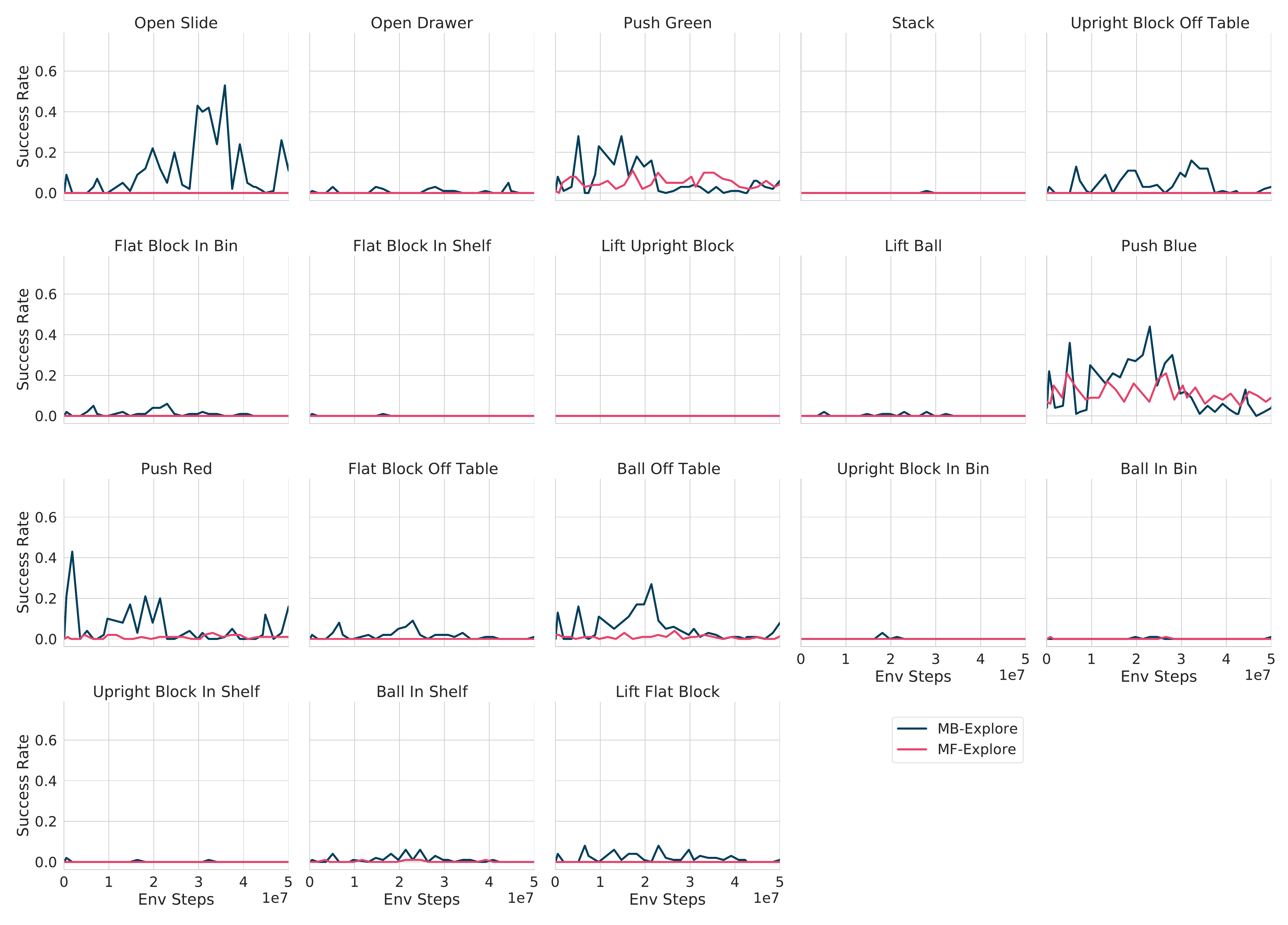}%
 \caption{Success rates on all 18 RoboDesk tasks during pre-training (reported for a single seed).}
 \label{fig:robodesk_pretrain}
\end{figure}

\Cref{fig:robodesk_pretrain} shows the success rate on all the 18 tasks of RoboDesk during pre-training. Unlike Crafter, we do not observe significant correlation between exploration and task rewards. 

\begin{figure}[h!]
\centering
\includegraphics[width=\textwidth]{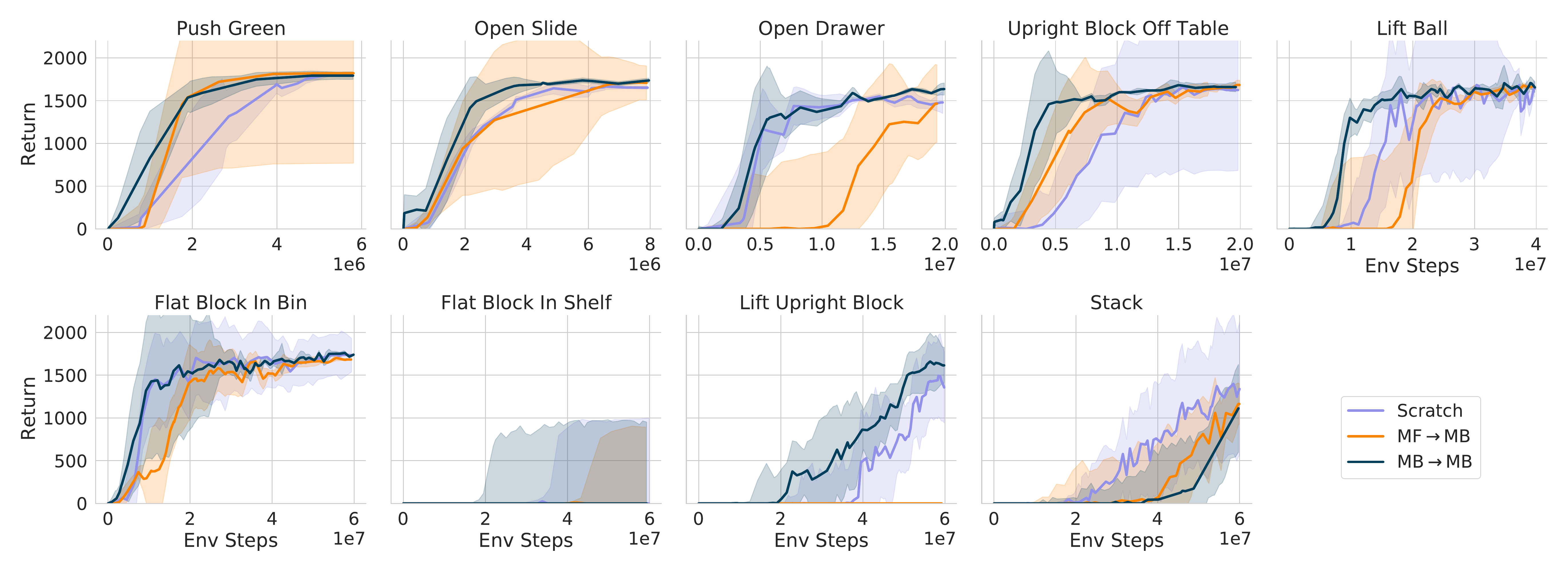}%
 \caption{Fine-tuning return curve on 9 RoboDesk core tasks. We take the median of 3 seeds and report the standard deviation.}
 \label{fig:robodesk_reward}
\end{figure}

In addition to the success rate reported in \Cref{fig:robodesk}, we report the return curve of model-based and model-free transfer agents, as well as the training from scratch agent, in \Cref{fig:robodesk_reward}. The relative comparison between agents judging from return is very similar to that from the success rate we presented in the main paper.

\clearpage
\subsection{Meta-World}

\begin{figure}[h!]
\centering
\subfloat[Average success rate on train tasks]{
\centering
\includegraphics[width=0.45\textwidth]{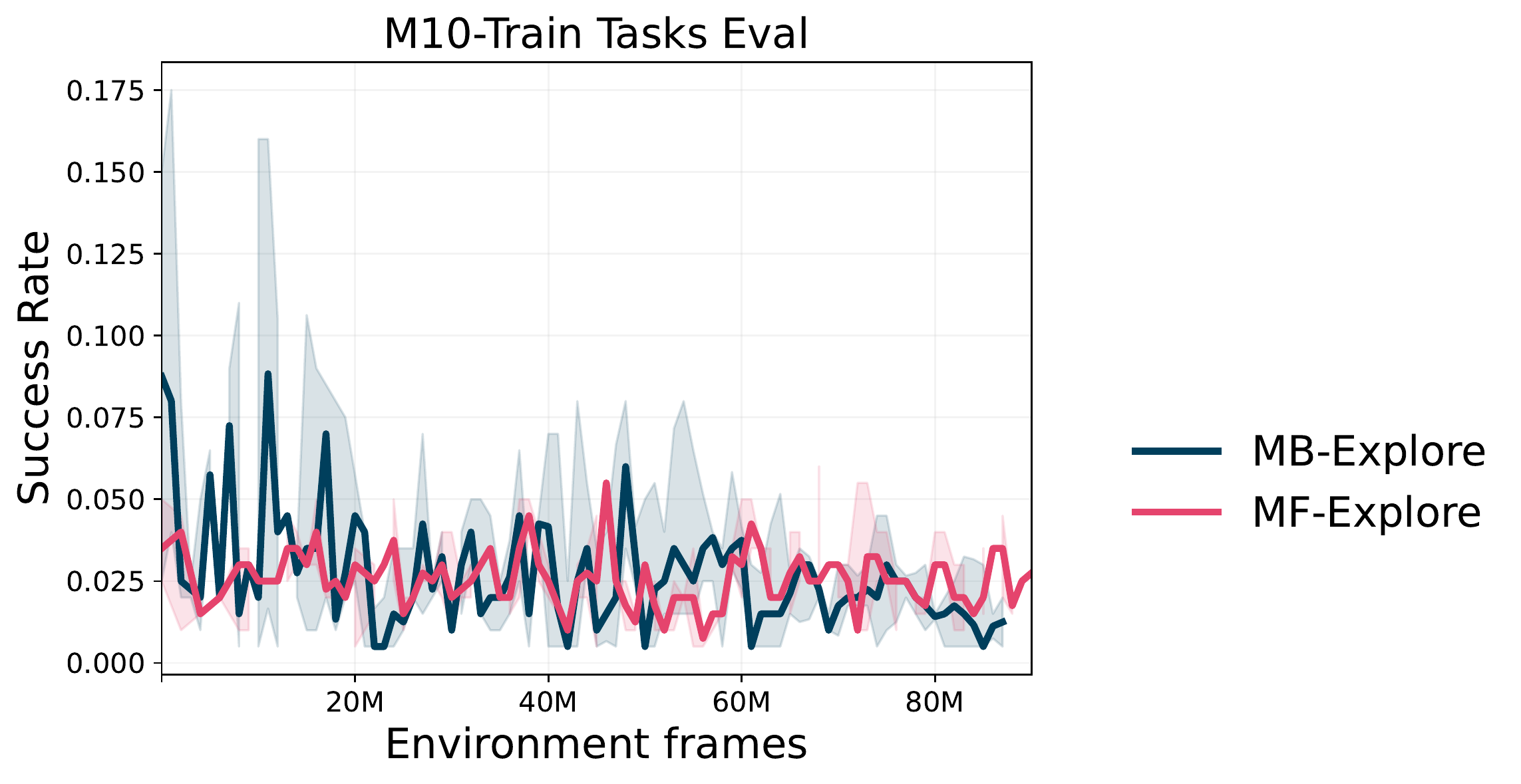}
\label{fig:metaworld_pretrain_success}
}\hfill
\subfloat[Average success rate on test tasks]{
\centering
\includegraphics[width=0.51\textwidth]{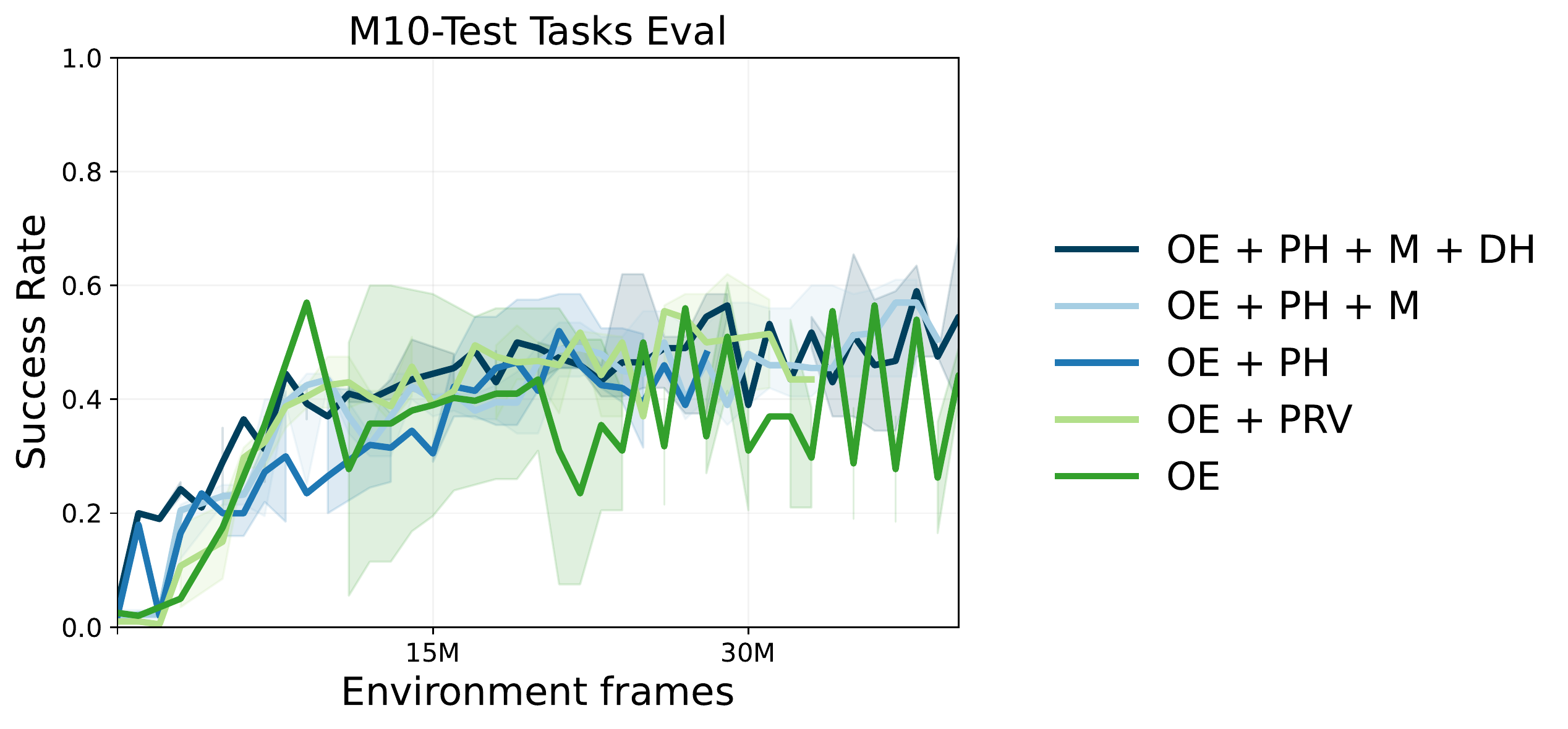}
\label{fig:metaworld_test_success_ablation}
}
\caption{Average success rate on Meta-World. \subref{fig:metaworld_pretrain_success} Average success rates on train tasks during pre-training phase. \subref{fig:metaworld_test_success_ablation} Average success rate on test tasks during fine-tuning phase for the ablation analysis.}
\end{figure}

Similar to RoboDesk, \Cref{fig:metaworld_pretrain_success} shows that there is no correlation between exploration and downstream tasks rewards. \Cref{fig:metaworld_test_success_ablation} provides the results for an ablation study on Meta-World. However, since there is no tangible benefits from the transfer in Meta-World, the various transfer setting results in similar performance.

\end{document}